\pdfoutput=1

\documentclass[11pt]{article}
\usepackage[dvipsnames]{xcolor}

 \usepackage{acl}

\usepackage{times}
\usepackage{latexsym}

\usepackage[T1]{fontenc}

\usepackage[utf8]{inputenc}
\usepackage{multirow}
\usepackage{microtype}
\usepackage{url}
\usepackage{graphicx}
\usepackage{wrapfig}
\usepackage{amsmath}
\usepackage{amssymb}
\usepackage{color,xcolor,colortbl}
\usepackage{enumitem}
\usepackage{longtable}

\usepackage{algorithm}
\usepackage{algpseudocode}
\usepackage{amsmath}

\usepackage{bm,bbm}
\usepackage{booktabs}
\usepackage{mathtools}
\usepackage{array}
\usepackage{subcaption}
\usepackage{pbox}
\usepackage{pifont}
\usepackage[scaled=0.7]{beramono}
\usepackage{tabularx}
\usepackage{tablefootnote}
\usepackage{float}
\usepackage{booktabs}
\usepackage{arydshln}
\title{JMLR: Joint Medical LLM and Retrieval Training for Enhancing Reasoning and Professional Question Answering Capability}

\author{%
  Junda Wang $^{1}$, 
  Zhichao Yang$^1$, 
  Zonghai Yao$^1$,
  Hong Yu$^{1, 2, 3, 4}$
  \\
    $^1$ Manning College of Information and Computer Sciences, University of Massachusetts Amherst, MA, USA\\
    $^2$ Department of Medicine, University of Massachusetts Medical School, Worcester, MA, USA\\
    $^3$ Miner School of Computer and Information Sciences, University of Massachusetts Lowell, MA, USA\\
    $^4$ Center for Healthcare Organization and Implementation Research, VA Bedford Health Care, MA, USA\\
  {\tt \{jundawang, zhichaoyang, zonghaiyao, hongyu\}@umass.edu}\\ 
  \\
}

\begin{document}

\maketitle

\begin{abstract}

Large Language Models (LLMs) have demonstrated a remarkable potential in medical knowledge acquisition and question-answering. However, LLMs can potentially hallucinate and yield factually incorrect outcomes, even with domain-specific pretraining. Previously, retrieval augmented generation (RAG) has limited success in addressing hallucinations. Unlike previous methods in RAG where the retrieval model was trained separately from the LLM, we introduce JMLR (for Jointly trains LLM and information Retrieval) during the fine-tuning phase. The synchronized training mechanism enhances JMLR's ability to retrieve clinical guidelines and leverage medical knowledge to reason and answer questions and reduces the demand for computational resources. We evaluated JMLR on the important medical question-answering application. Our experimental results demonstrate that JMLR-13B (70.5\%) outperforms a previous state-of-the-art open-source model using conventional pre-training and fine-tuning Meditron-70B (68.9\%) and Llama2-13B with RAG (67.7\%) on a medical question-answering dataset. 
Comprehensive evaluations reveal JMLR-13B enhances reasoning quality and reduces hallucinations better than Claude3-Opus. Additionally, JMLR-13B (148 GPU hours) also trains much faster than Meditron-70B (42630 GPU hours).  Through this work, we provide a new and efficient knowledge enhancement method for healthcare, demonstrating the potential of integrating retrieval and LLM training for medical question-answering systems.

~\footnote{The code(https://github.com/believewhat/JMLR-Joint-Medical-LLM-and-Retrieval-Training), along with selected retrieval data that can be made public, is included in the supplementary material and will be made publicly accessible with CC-BY 4.0 license upon the paper's acceptance.}.

\end{abstract}

\section{Introduction}

\begin{figure}
    \centering
\includegraphics[width=\linewidth]{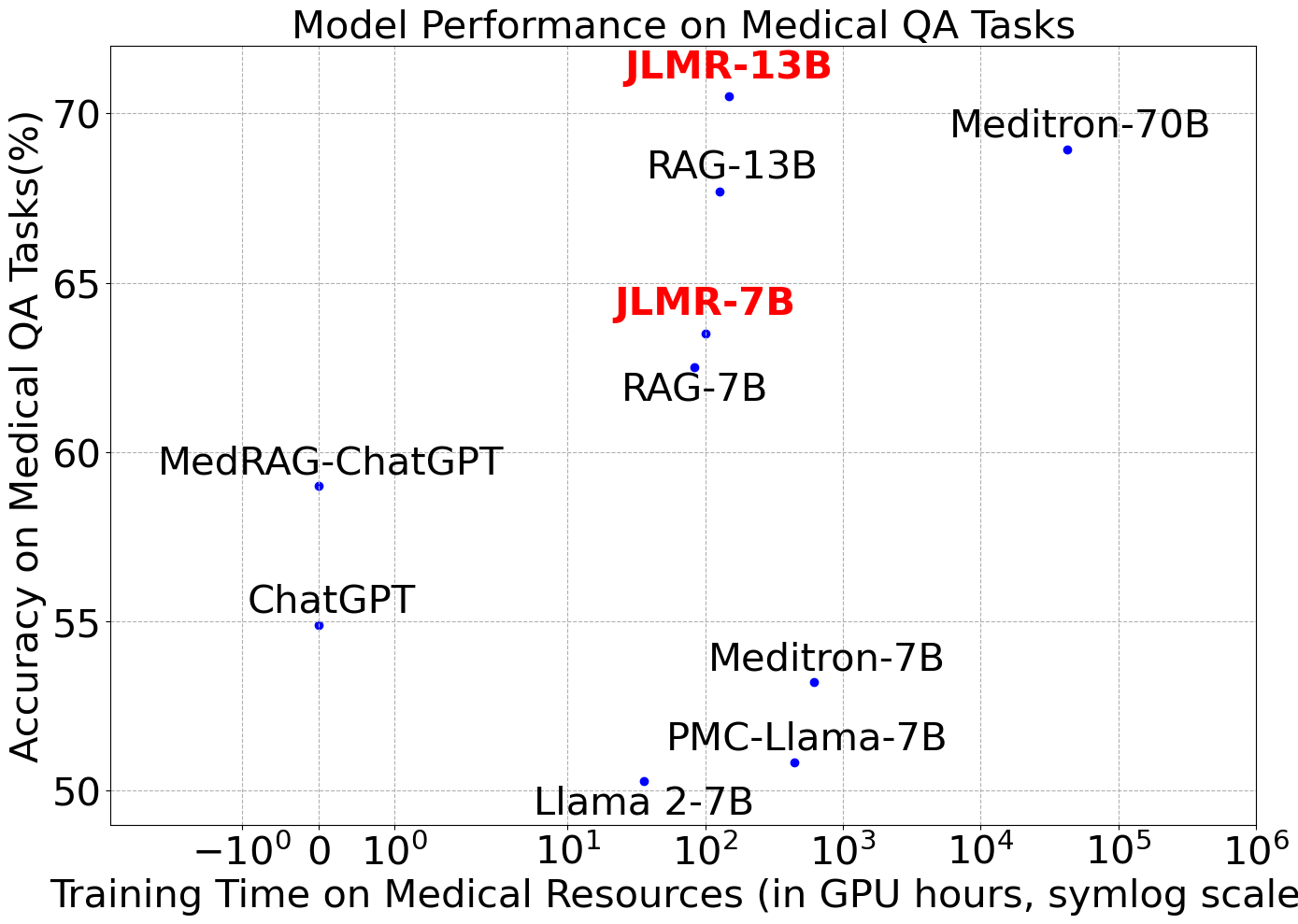}
        \vspace{-4mm}
        \caption{JMLR achieved the highest average accuracy across the MMLU-Medical, MedMcQA, MedQA, and Amboss datasets, utilizing only 148 GPU hours.}
        \label{fig:tradeoff}
        \vspace{-4mm}
\end{figure}

\begin{figure*}
    \centering
\includegraphics[width=0.9\textwidth]{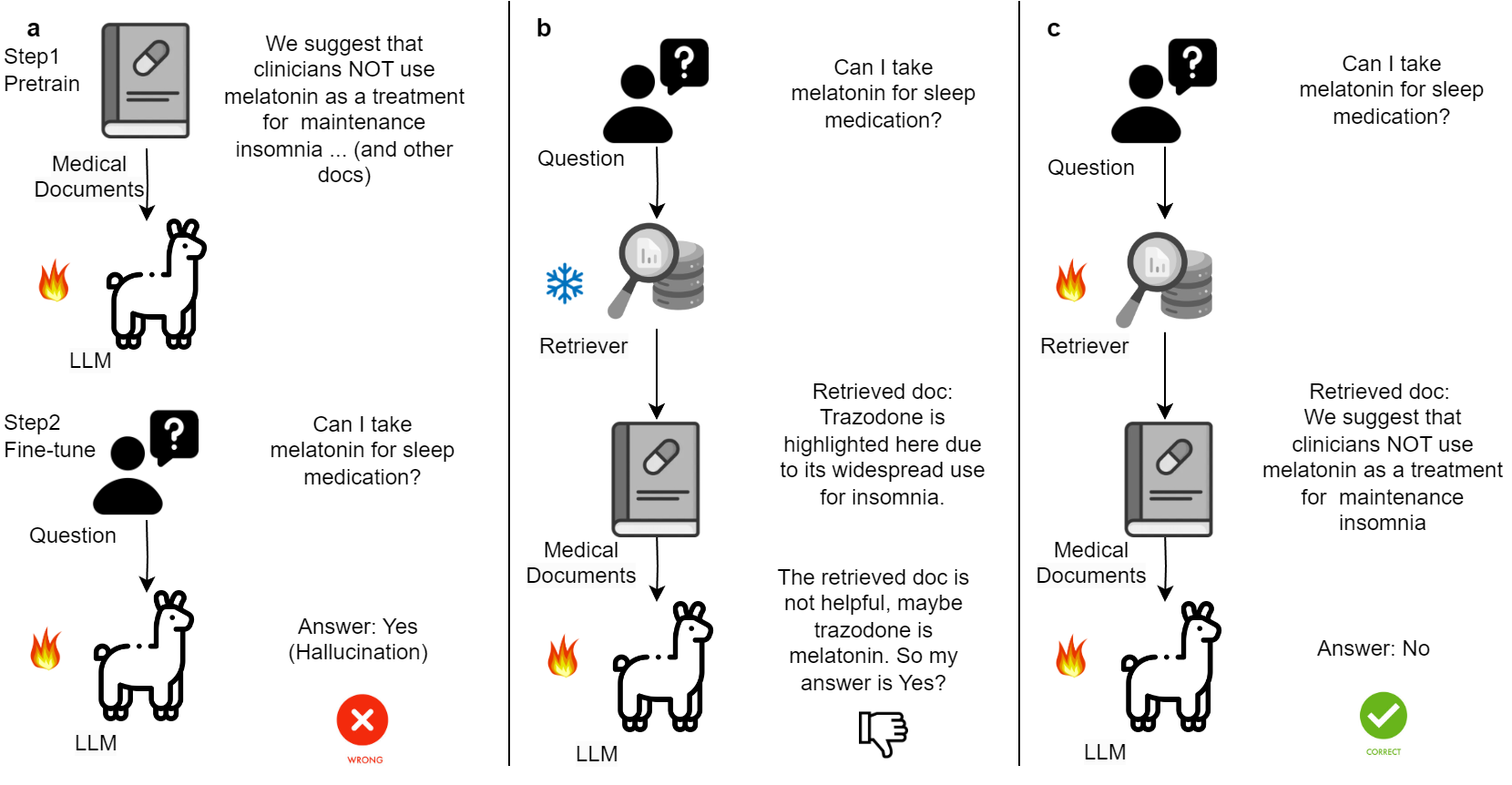}
        \vspace{-2mm}
        \caption{Comparison between different domain adaptation methods: traditional domain pretraining method (left), RAG (middle), and JMLR (right). JMLR retrieves the documents to reduce the hallucination. Parameters are updated simultaneously for the retriever and large language models (LLM) models, leading the retriever to know which domain-specific document is helpful for LLM to give a reasonable answer.}
        \label{fig:sample}
        \vspace{-4mm}
\end{figure*}


Effective clinical decision-making relies on a logical diagnostic chain, which requires specialized knowledge that isn't widely accessible, presenting a major healthcare challenge. Large Language Models (LLMs) show promise in making this critical medical knowledge more accessible~\citep{peng2023study,yang2023performance}.
Traditionally, LLMs have been developed for general tasks using data from diverse online sources, leading to a lack of high-quality, domain-specific information, especially in medicine~\cite {Wornow2023TheSF,singhal2023large}. 
Previous researches expand medical knowledge by continuing training general domain LLMs with domain-specific datasets.~\citep{yunxiang2023chatdoctor,zhang2023huatuogpt,toma2023clinical}.

However, LLMs face the significant challenge of ``hallucination'', where models generate plausible but incorrect or unverified information~\citep{ji2023survey,bang2023multitask}. Such errors raise serious concerns in healthcare, where accuracy is imperative~\citep{ahmad2023creating}. 
Hence, recent works proposed retrieval augmented generation (RAG): they first train a retriever to obtain relevant documents from a general domain corpus based on the input query and then train an LLM to generate a response based on the input query and the retrieved documents ~\citep{lewis2020retrieval, pmlr-v162-borgeaud22a, Cheng2023LiftYU, Hiesinger2023AlmanacRL, Xiong2024BenchmarkingRG}. 
By retrieving domain-specific documents, RAG identifies relevant knowledge and provides contextual grounding for LLMs, thereby alleviating hallucination issues ~\citep{shuster-etal-2021-retrieval-augmentation, zhu2023large}. 

Although RAG shows high accuracy in benchmarks for open-domain question answering, its effectiveness in specific domains is yet in challenge because retrievers trained in the general domain usually perform worse than those fine-tuned in the specific domain ~\citep{Gao2023RetrievalAugmentedGF,Zhao2022ALD,Thakur2021BEIRAH}.
Fine-tuning retrievers requires document-query pairs, which may not be readily available for a specific domain. 
Labeling such medical pairs specifically for this purpose incurs additional time and financial costs. 
Moreover, RAG trains retriever and LLM separately.
When training LLM with a frozen retriever, 
the retrieval may not be optimized to align with LLM to produce the correct answer \citep{asai2024reliable, Yoran2023MakingRL, rubin-etal-2022-learning}. 
In this study, we introduce a novel method to jointly train retrieval and LLM. 
Although the method is generalizable, we focus on evaluating it in the important medical domain, where the accuracy of the generated content is crucial.

Specifically, we introduce Joint Medical LLM and Retrieval Training (JMLR), a novel approach that jointly train LLM and retriever. 
As shown in Figure~\ref{fig:sample}, JMLR presents a novel approach compared to the traditional pretrain-finetune process. 
JMLR enhances question answering by fetching relevant domain-specific documents that maximize the potential of LLMs. 
This involves including each retrieved document in the initial input question before the LLM's response attempt, thereby augmenting the input to train the LLM for answer generation.
For training jointly, JMLR introduces a unique mechanism called LLM-Rank loss to train the retriever. 
This is achieved by evaluating the improvement in LLM's performance upon including any candidate retrieved documents.
We calculate the log probability of the LLM's answer with each candidate, adopting the negative of these values as the relevance score for each candidate. 
When reducing this loss, the retriever is trained to prioritize candidate documents that significantly aid the LLM. 
Our results show that the joint training improves medical question answering, especially in scenarios requiring nuanced understanding and specific information retrieval.


To validate JMLR's effectiveness in reducing computational resource requirements and fully utilizing given knowledge, we designed three experimental setups. These experiments evaluated whether direct fine-tuning, instead of traditional pretraining, would improve JMLR's performance.
We utilized external data resources for retrieval, including the MIMIC-IV dataset~\citep{johnson2020mimic}, medical textbooks, and diverse medical documents for knowledge expansion.
We experimented with different LLMs, including PMC-Llama, Meditron, GPT-3.5 with or without MedRAG.  Our JMLR-7B has demonstrated superior results, achieving an impressive 62.3\% accuracy, outperforming the traditional method's 53.2\% on the medical QA dataset. 
Furthermore, our 13B model (70.5\%) surpasses the performance of both the open-source medical LLM (Meditron 70B: 68.9\%) and closed-source general LLM (GPT-3.5: 54.9\%). 
Moreover, evaluation by both GPT-4 and domain experts supports the superior performance of JMLR. 
We summarize our key contributions as follows:

\begin{itemize}
[leftmargin=.2in,topsep=2pt]
\setlength\itemsep{0.01em}
\vspace{-0.2em}
    \item We propose a novel method that jointly trains retrieval and LLMs, resulting in JMLR-13B surpassing the state-of-the-art 70B open-source model in several medical question-answering benchmarks. 
    \item Through comprehensive automated and human evaluations, we demonstrate that JMLR-13B consistently enhances models' reasoning quality while significantly reducing hallucinations, achieving superior reasoning ability compared to Claude3-Opus.
    \item JMLR shows promise to enhance domain-specific retrieval by fully eliminating additional human annotation.
    \item When training 7B parameter model with a single Nvidia A100 GPU, our training method decreases the training time required for knowledge expansion to just 100 hours. This is a considerable reduction compared to Meditron's pretrain-finetune process, where pretraining alone takes 588 hours, and finetuning adds an additional 36 hours.
\end{itemize}

\setlength{\belowdisplayskip}{4pt} \setlength{\belowdisplayshortskip}{4pt}
\setlength{\abovedisplayskip}{4pt} \setlength{\abovedisplayshortskip}{4pt}


\section{Problem Formulation \& Traditional Method}

Given a set of medical questions $\mathcal{Q}$ and a set of medical documents $\mathcal{D}$ containing medical knowledge, our goal is to construct a language model $\mathcal{M}$ that can provide accurate answers.

Formally, for each question $q_i \in \mathcal{Q}$, there is a corresponding correct answer $a_i^*$ within a set of options $A_i$, where $a_i^* \in A_i$. The model $\mathcal{M}$ maps each question to a predicted answer:
\begin{equation*}
\hat{a}_i = \mathcal{M}(q_i, \mathcal{D}; \theta)
\end{equation*}
where $\theta$ represents the parameters of the model.

Our objective is to find the optimal parameters $\theta^*$ that minimize the loss function $\mathcal{L}$, which measures the discrepancy between the predicted answer $\hat{a}_i$ and the correct answer $a_i^*$:
\begin{equation*}
\theta^* = \arg\min_\theta \sum_{i=1}^{|\mathcal{Q}|} \mathcal{L}(\hat{a}_i, a_i^*)
\end{equation*}
The loss function can be instantiated as a cross-entropy loss for classification tasks, where the number of classes is equal to the number of options.

To solve this problem, previous methods pretrained LLM to learn medical knowledge for a medical LLM, and then finetuned medical LLM on medical QA task~\citep{chen2023meditron,wu2023pmc,yunxiang2023chatdoctor,toma2023clinical}. 
Specifically, such traditional pretrain-finetune pipeline first continued pretraining general domain LLMs on medical documents $\mathcal{D}$ with next-token prediction loss function, and then finetuned the medical LLMs to select $a_i^*$ given question $q_i$ and option description with loss function $\mathcal{L}$.

\section{Joint Medical LLM and Retrieval Training (JMLR)}

In comparison to traditional method, JMLR did not pretrain LLM on medical documents. Instead, JMLR selected question-related medical documents as additional context during fine-tuning of an open-domain LLM. Specifically, given a question $q_i$, we trained retriever to find helpful medical documents from $D$, which was then used to train LLM to generate the final answer $a_i^*$.

\subsection{Retriever} 

For information retrieval, we used the ColBERT model and the trained weight as the initial model. ColBERT utilizes BERT-based encoders to transform both queries and documents into bags of embeddings. A single BERT model is shared between the query and document encoders, but the inputs are distinguished by appending special tokens: $[Q]$ for queries and $[D]$ for documents. The text is tokenized for queries into BERT-based WordPiece tokens $q_0, q_1 \ldots q_l$, with the $[Q]$ token placed after BERT’s sequence-start token $[CLS]$. For documents, ColBERT segments a document d into its constituent tokens $d_1, d_2, \ldots d_n$, to which we prepend BERT’s start token $[CLS]$ followed by our special token $[D]$ that indicates a document sequence. Queries shorter than a predefined number of tokens $N_q$ are padded with BERT's special $[mask]$ tokens to reach the length $N_q$, otherwise truncated. The padded sequence of input tokens is then passed through BERT. The resulting embeddings for queries ($E_q$) and documents ($E_d$) are computed as follows: $E_q$ is normalized after being processed through a CNN layer applied to BERT’s output, and $E_d$ undergoes a similar process but includes a filtering step to remove embeddings of punctuations.

$
\begin{aligned}
& E_q:=\operatorname{CNN}\left(\operatorname{BERT}\left([Q] q_0 q_1 \ldots q_l\right)\right) \\
& E_d:=\operatorname{Filter}\left(\operatorname {CNN} \left(\operatorname {BERT} \left([D] d_0 d_1 \ldots d_n\right)\right)\right)
\end{aligned}
$

\begin{equation*}
S_{q, d}:=\sum_{i \in\left[\left|E_q\right|\right]} \max _{j \in\left[\left|E_d\right|\right]} E_{q_i} \cdot E_{d_j}^T
\end{equation*}

\subsection{LLM}

We adapted Llama as our LLM model. The original Llama model supports only a maximum of 4k tokens, which poses a challenge when multiple documents are retrieved.
To address this, we utilized the Shifted Sparse Attention (S2-Attn) mechanism~\citep{chen2023longlora}, which can be used to process long sequences by 
mitigating the high memory cost and slow processing time associated with standard self-attention in LLMs. S2-Attn divides input into sequence spans within self-attention modules and introduces shifted patterns for inter-span information exchange. This extends context length efficiently without extra computational costs and achieves near-baseline performance.

\subsection{JMLR Architecture} 

However, Colbert matches queries and documents based on the retriever similarity, returning scores and documents. To enable the retriever system to return documents that are more helpful for providing accurate answers to Llama, we constructed the JMLR architecture. Let $\mathcal{Q}$ be a set of queries, $\mathcal{D}$ be a corpus of domain-specific documents, and $\mathcal{A}$ be the set of options. Our task is to construct a function $f$ parameterized by $\theta$, which maps a question $Q \in \mathcal{Q}$ and a set of documents $D \subseteq \mathcal{D}$ to an answer $A \in \mathcal{A}$.

The objective is to simultaneously learn the optimal parameters $\theta^*$ and $\phi^*$ that minimize the combined loss on the retrieved documents and the generated answers, formalized as:
\scalebox{0.82}{
$
(\theta^*, \phi^*) = \arg\min_{\theta, \phi} \mathbb{E}_{(Q,A) \sim \mathcal{T}} \left[ \mathcal{L}(f_A(Q, f_R(Q; \phi), \theta), A) \right]
$
} 
where $\mathcal{L}$ is a loss function that evaluates the correctness of the answer and the relevance of the retrieved documents. The retrieval function $f_R$ is defined by:
$$
f_R(Q; \phi) = \arg\top_{D \subseteq \mathcal{D}, |D|=k} \text{rel}(Q, D; \phi)
$$
and the similarity function $\text{rel}(Q, D; \phi)$ scores how relevant each document $D$ is to the question $Q$.

The combined optimization involves updating both $\theta$ and $\phi$ through gradient descent to minimize the expected loss:
$$
\nabla(\theta, \phi) \propto \nabla_{\theta, \phi} \mathcal{L}(f_A(Q, f_R(Q; \phi), \theta), A)
$$
where $\nabla_{\theta, \phi} \mathcal{L}$ denotes the gradient of the loss function with respect to both sets of parameters. This dual-parameter optimization ensures that the retrieval function is aligned with the needs of the answer function, enhancing both the accuracy of the responses and the relevance of the information retrieved. To solve this, JMLR structure employs a rank loss, $L_{\text{rank}}$, which links the Llama and retriever systems. Llama's content quality influences the updating of retriever parameters. If the returned document $i$ reduces Llama's loss function $\hat{L}$ more than document $j$, it indicates that document $i$ is more helpful for answering than document $j$. In calculating the rank loss, we incorporate the scores generated by the retriever system. However, due to the large variance in scores, we normalize them to obtain $\hat{S}$. The essence of $L_{\text{rank}}$ is that the retriever system's score rankings should be updated based on the ranking.
in $\hat{L}$.
\[
\mathcal{L} = L_{\text{rank}} + L(f_A(Q, f_R(Q; \phi)), \theta, A),
\]
\[
L_{\text{rank}} := -\frac{1}{M} \sum_{i=1}^M \mathbb{I}(L_i \neq 0) \cdot F_i,
\]
where M is the number of queries in the training set and \(F_i\) is calculated as:
\[
F_i = \mathbb{I}(L_i) \log(C_i) + (1 - \mathbb{I}(L_i)) \log(1 - C_i),
\]
and \(C_i\) represents the contrastive logits:
\[
C_i = \sigma(\hat{S}_{q_i, d_{k}}) - \sigma(\hat{S}_{q_i, d_{j}}),
\]
with the associated decision-based loss differences:
\[
L_i = \hat{L}_{i}(y, y'(x, d_j)) - \hat{L}_{j}(y, y'(x, d_k)).
\]

For JMLR model training, the dataset comprises QA pairs, with the input being a question and the output being the correct choice, including the rationale part. The top 7 documents with the highest scores are aggregated and fed into the model during the training process. Prior to training, we extracted the top 30 documents based on scores from the retriever. During each iteration, 30 scores are generated by the retriever, and we perform weighted random sampling based on these scores, with higher scores having a greater probability of selection. This approach ensures that each document inputted into the LLM model is dynamic. The LLM will likely encounter different useful documents and some that are not useful. This step enhances the LLM's robustness, helping it discern which knowledge from which documents is useful and which documents are irrelevant.

\section{Experiment}

\subsection{Dataset}
\paragraph{Medical document:}  
Data quality of medical documents is important for LLMs to answer medical questions. 
Previous work primarily sourced from a wide array of medical research articles and clinical guidelines. 
Research papers, like those found in PubMed, provide foundational and current information on healthcare.
Clinical guidelines are thoroughly designed protocols developed to assist clinicians in making decisions given patient medical history.
Other high-quality medical documents to retrieve include medical textbooks.

\begin{table}[H]
\centering
\scalebox{0.8}{
\begin{tabular}{lccc}
\hline
Corpus      & \#Doc.      \\ \hline
PubMed      & 23.9M           \\
Textbooks   & 18           \\
Cancer Care Ontario & 87  \\
Center for Disease Control and Prevention & 621  \\
Canadian Medical Association & 431  \\
International Committee of the Red Cross & 49  \\
National Institute for Health and Care Excellence & 1.7k  \\
Strategy for Patient-Oriented Research & 217  \\
World Health Organization & 223  \\
WikiDoc & 33k  \\ \hline
\end{tabular}
}
\vspace{-2mm}
\caption{For each medical corpus source, we provide the number of distinct documents, the approximate articles across all documents}
\vspace{-4mm}
\label{tab:my_label}
\end{table}

\paragraph{Medical QA:} 

The MedQA dataset~\cite{jin2021disease}, sourced directly from the official USMLE website, included a range of sample questions for Step1, Step2CK, and Step3, released between June 2022 and March 2023. These questions represent the complex medical knowledge and ethical scenarios that medical students and practitioners are expected to navigate. 

Additionally, Amboss question bank is a comprehensive resource widely used by medical professionals and students. This dataset provided an extensive array of Step1, Step2CK, and Step3-type questions, further enriching our training material with practical and diverse medical scenarios. 

The MedMCQA dataset comprises 194k multiple-choice questions from Indian medical entrance exams, covering 2.4k healthcare topics and 21 medical subjects~\citep{pal2022medmcqa}.

MMLU-Medical is selected from the MMLU dataset~\cite{hendrycks2020measuring}, focused on nine subjects most pertinent to medical and clinical knowledge — high school biology, college biology, college medicine, professional medicine, medical genetics, virology, clinical knowledge, nutrition, and anatomy. Since this dataset has no training set, we opt to evaluate it using the LLM originally trained on MedMCQA.

We individually apply supervised finetuning on each QA dataset and subsequently assess their performance on the corresponding test sets, unless otherwise specified. MedQA and Amboss both offer detailed rationale with long explanations.

\subsection{Baselines}
For comparison with traditional domain pretraining method, we used Meditron, an open source LLM pretrained on clinical guidelines and research papers \citep{chen2023meditron}.
We also compared with other closed source LLMs such as OpenAI GPT-3.5, GPT-4~\cite{achiam2023gpt}, and Anthropic-Claude3 ~\cite{anthropic2024claude}
For the rationale evaluation experiment, we didn't include Meditron because it cannot generate a valid rationale in our experiments. 
We didn't include GPT-4 because it was used as the evaluator. 
To validate the benefit of joint training, we also compared JMLR with its naive version: RAG, which freezes the retriever and fetches the same document given a question.

\subsection{Evaluation Metrics}
For automated metrics, we used accuracy to evaluate the model's ability to select the correct final choice. 
We evaluated the rationale quality using metrics UMLS-F and GPT-4 score. 
UMLS-F is a factuality metric based on F1 score between entities in gold explanation and generated explanation~\footnote{More UMLS-F details can be found in Appendix~\ref{apx:umls-f}}. 
GPT-4 score is a reference-free metric.
GPT-4 was asked to rate the generated explanation on a Likert scale from 1 to 5 (higher better) along the 3 dimensions: question comprehension (e.g., indication the question has been understood), recall of knowledge (e.g., mention of a relevant and/or correct fact for answering the question), and medical reasoning (e.g., correct reasoning for answering the question) \citep{singhal2023large}. We put GPT-4-score prompts in Table~\ref{gpt4-score-prompt}.

For human evaluation, we engaged three medical professionals to review and assess a small sample of 20 questions drawn from the test sets. These doctors were asked to provide their preference for the rationales generated by two models (GPT-3.5 and JMLR-13b) given the gold rationale.

To validate the reliability of the selected automated metrics, we calculated the Cohen's Kappa between UMLS-F and expert preference, as well as Kappa between the GPT-4 score and expert preference, obtaining results of 0.69 and 0.81, respectively. These results indicate that both metrics exhibit substantial to almost perfect agreement with human evaluation.


\subsection{Training Details}

Both our training process and the conventional finetuning approach employ the AdamW optimizer, with $\beta_1 = 0.9$, $\beta_2 = 0.95$, and $\text{eps} = 1 \times 10^{-5}$. We implement a cosine learning rate schedule, incorporating a warmup phase that accounts for 10\% of the training duration and decays the learning rate to 10\% of its peak value. In alignment with the practices outlined in Llama 2-chat \cite{Touvron2023Llama2O}, our training employs a learning rate of $1 \times 10^{-5}$, a weight decay factor of 0.1, and manages a batch size of 2. The finetuning phase spans 5 epochs for all iterations. However, we apply a distinct learning rate for ColBERT, set at $3e-5$. The optimization strategy for training ColBERT mirrors that used for Llama 2. Throughout these experiments, we utilize four A100 80 GB GPUs.

\section{Main Results}

As shown in Table \ref{sota}, our JMLR with 13B parameters outperforms previous SOTA open access model Meditron 70B parameters and even closed access model such as GPT-3.5.

\begin{table}
\centering
\scalebox{0.75}{
\begin{tabular}{l|cccc|c}
\hline
Model & \footnotesize{MedQA} & \footnotesize{Amboss} & \footnotesize{MMLU-M} & \footnotesize{MedMCQA} & \footnotesize{AVG} \\
\hline
GPT-3.5 &50.2 &49.1 &69.4 &51.0 &54.9 \\
GPT-4 &74.7 &82.1 &88.4 &69.5 &78.6 \\
\hline
\footnotesize{Meditron-70B} &60.7 &76.4 &73.6 &65.1 &68.9 \\
RAG-13B &59.9 &76.9 &69.9 &64.2 &67.7 \\
JMLR-13B &62.5 &81.2 &72.8 &65.5 &70.5 \\
\hline
\end{tabular}
}
\vspace{-2mm}
\caption{Comparison between JMLR and SOTA open (Meditron, RAG, JMLR) or closed (GPT-3.5/4) LLMs.}
\vspace{-4mm}
\label{sota}
\end{table}

\subsection{Domain Retrieval > Domain Pretrain}

Domain retrieval models (both RAG-7B and our JMLR-7B) outperform domain-pretrained LLMs across all medical QA benchmarks, as shown in Table~\ref{benchmark}. 
In the MedQA and Amboss datasets, where JMLR-7B not only outperforms baselines but does so with a notable margin, scoring 51.3\% in MedQA and 68.3\% in Amboss. This is particularly significant when compared to Meditron-7B, which scores 47.9\% and 50.1\% in these datasets, respectively. The trend continues in MMLU-Medical and MedMcQA datasets, JMLR-7B achieved scores of 65.3\% and 64.1\%, surpassing the scores of Meditron-7B (55.6\% and 59.2\%). 
Overall, the results illustrate that JMLR-7B model, on average, achieves about a 17\% improvement in performance over its closest competitor, Meditron-7B.
This highlights the effectiveness of our training method and the substantial advancements it brings to the field of medical benchmarking.

The enhanced performance of JMLR-7B can be partially attributed to its use of Llama-2 as the foundational model, showcasing significantly higher average performance than other pre-trained baselines. 
However, the distinct edge comes from our integrated training approach that combines retriever and LLM. 
This methodology not only further enhances Llama-2's performance in medical benchmarks but also equips the retriever component with the ability to effectively source relevant and beneficial documents to aid the LLM in answering questions.
In contrast, GPT-3.5, when devoid of medical guidelines, has been known to make basic mistakes, such as recommending vaccinations to pregnant women—a practice that contradicts medical norms as documented in Table~\ref{tab:case1}. 
Our model, with the support of guidelines, avoids such errors. 
This synergistic approach substantially reduces hallucinations and bolsters the model's overall reliability.

\begin{table}
\centering
\scalebox{0.68}{
\begin{tabular}{l|cccc|c}
\hline
Model &MedQA &Amboss &MMLU-M &MedMCQA &AVG \\\hline
Llama 2 &44.0 &46.5 &56.3 &54.4 &50.3 \\
Meditron* &47.9 &50.1 &55.6 &59.2 &53.2 \\\hline
RAG\# &47.3 &50.7 &63.8 &62.1 &55.9 \\
JMLR\# &\textbf{51.3} &\textbf{68.3} &\textbf{65.3} &\textbf{64.1} &\textbf{62.3} \\
\hline
\end{tabular}
}
\caption{Comparison between domain retrieval (\#) and domain pretraining (*) using models of the same 7B size and domain data (Open Guidelines).}
\vspace{-4mm}
\label{benchmark}
\end{table}

\subsection{Joint Training > RAG}


During fine-tuning, synchronously updating both the Retriever and LLM yields better results than updating only the LLM. Table \ref{sota} shows that JMLR models significantly outperform traditional RAG methods. For example, JMLR-13B's scores (72.8\% on MMLU, 65.5\% on MedMcQA) is far superior to RAG-13B (69.9\% on MMLU, 64.2\% on MedMcQA). This result suggests that without updating the retriever, the documents retrieved may be relevant to the question but not necessarily helpful to the LLM's response. 
Joint training ensures that the retriever learns which documents are beneficial for LLM.
To evaluate the adaptability of the JMLR retriever, we further applied it to other LLMs. 
Specifically, we employed different retrievers to identify relevant documents, which were then provided to GPT-3.5 for generating the final answer. 
The retriever trained with JMLR attained an average accuracy of 57.5\%, while the retriever from RAG achieved an accuracy of 56.7\%. 
In addition, we present a case study in Table~\ref{tab:case1}, where JMLR retrieves closely related cases.
These results indicate that the JMLR retriever effectively identifies documents that not only support Llama but also aid GPT-3.5 in selecting the correct answers.

\subsection{Enhancing Model Rationale Quality}

\begin{table}[ht]
\vspace{-2mm}
    \centering
    \scalebox{0.7}{
    \begin{tabular}{lcccc}
        \toprule
        & \textbf{Haiku} & \textbf{Opus} & \textbf{GPT3.5} & \textbf{JMLR} \\
        \midrule
        \textbf{UMLS Factuality} & 0.2337 & 0.2356 & 0.2187 & \textcolor{red}{0.2463} \\
        \textbf{GPT-4 Overall} & 4.0559 & 4.2449 & 4.0620 & \textcolor{red}{4.3036} \\
        \hline
        \textbf{GPT-4 Comprehension} & 4.7107 & 4.7561 & 4.7549 & 4.7661 \\
        \textbf{GPT-4 Reasoning} & 3.6701 & 4.0976 & 3.6863 & 4.0101 \\
        \textbf{GPT-4 Recall} & 3.6726 & 3.8780 & 3.6849 & 4.1661 \\
        \bottomrule
    \end{tabular}
    }
    \caption{UMLS-F and GPT-4 score across 4 different models (e.g., Claude3, GPT3.5, JMLR-13B).}
    \label{tab:performance_scores}
    \vspace{-4mm}
\end{table}
In addition to significantly improving the model's performance in QA accuracy, JMLR also plays a crucial role in helping the model generate higher-quality rationales with fewer hallucinations.
As shown in the Table~\ref{tab:performance_scores}, both UMLS Factuality and GPT-4 Overall scores show that JMLR-13B outperforms the other three models, demonstrating its high-quality rational generation.
This result aligns with our expert evaluations on the small samples, where JMLR-13B achieved a win rate of 0.60 compared to GPT-3.5.

Specifically, GPT-4 Comprehension primarily measures the plausibility of generated reasons. As shown in Table \ref{tab:performance_scores}, the scores for GPT-4 Comprehension are uniformly high across all models, with only minor differences, indicating that all these models can generate logically coherent explanations for the questions.
GPT-4 Reasoning evaluates medical logical reasoning abilities. JMLR-13B stands out in this critiria because its retrieval guidelines include valuable content, such as information on Differential Diagnosis, which aids the model in more effectively associating key information like symptoms, diseases, and medications presented in the questions.
Finally, GPT-4 Recall assesses factuality, i.e., whether the model can hit the critical information points. 
Our results demonstrate that JMLR-13B achieves the most significant improvement in this criteria.
The high-quality documents retrieved by JMLR-13B provide the necessary key information, allowing the model to generate higher-quality rationales around these key points. 
The assessments of UMLS Factuality are similar to GPT-4 Recall, and their final results are consistent.
We further explore this in the appendix~\ref{sec:appendix} and provide case studies to support our findings.




\section{Ablation Study}

\begin{table}[h!]
\vspace{-4mm}
\centering
\scalebox{0.8}{
\begin{tabular}{lcccc}
\toprule
& MedQA & Amboss \\
\midrule
FT-Llama-7B & 44.0 & 46.5 \\
FT-Llama-7B-ColBERT & 40.6 & 45.6 \\
JMLR-7B-Separate & 55.3 & 69.0 \\
JMLR-7B & 56.2 & 71.2 \\
\bottomrule
\end{tabular}
}
\vspace{-2mm}
\caption{We conducted an ablation study to verify the necessity of the JMLR training method. We discussed retrieval versus no retrieval and asynchronous training versus joint training. Ultimately, we found that the JMLR training method significantly outperformed the other approaches.}
\vspace{-4mm}
\label{tab:model_performance}
\end{table}

To assess the effects of fine-tuning and information retrieval on QA accuracy, we compare \textit{JMLR-7B} (on medical documents) with the following new baselines: 
The \textit{JMLR-7B-Separate} model updates the parameters of both Llama and the retriever similar to JMLR-7B, but in an asynchronous manner. Initially, it fixes the parameters of the retriever while only updating those of Llama. Subsequently, it reverses this process, updating only the retriever's parameters using rank loss. 
The \textit{FT-Llama-7B+ColBERT} model, on the other hand, freezes the retriever in a manner akin to RAG-7B. However, the retriever in this setup is solely applied during inference and not during fine-tuning. 
The \textit{FT-Llama-7B} serves as a naive fine-tuned baseline where the retriever is not employed. We fine-tuned the Llama 2-7B on the training sets. 
The accuracy are shown in Table~\ref{tab:model_performance}. 
From the table, we can observe that:

\begin{itemize}
[leftmargin=.2in,topsep=2pt]
\setlength\itemsep{0.01em}
\vspace{-0.2em}
    \item \textit{FT-Llama-7B+ColBERT} (44.0\% on MedQA, 46.5\% on Amboss) showed significant improvements over \textit{FT-Llama-7B}, indicating that the fine-tuning phase notably enhanced Llama’s capabilities in medical QA. However, this phase did not effectively leverage medical guidelines.
    
    \item \textit{JMLR-7B-Separate} model achieved superior performance (55.3\% on MedQA and 69.0\% on Amboss), surpassing the \textit{FT-Llama-7B-ColBERT} (40.6\% on MedQA, 45.6\% on Amboss). The distinctive fine-tuning approach of \textit{JMLR-7B-Separate}, which integrates medical documents during the tuning phase, enables the model to utilize medical resources more effectively.
    
    \item Finally, \textit{JMLR-7B} (56.2\% on MedQA, 71.2\% on Amboss) uses a joint training approach that facilitates the retrieval of highly relevant medical documents, surpassing the \textit{JMLR-7B-Seperate} (55.3\% on MedQA, 69.0\% on Amboss). The result demonstrate that jointly training Llama and ColBERT outperforms training the two separately.
\end{itemize}

\section{Related Work}
\paragraph{Medical Large Language Model} 
Medical LLMs have advanced from adapting models like BERT with biomedical datasets \citep{gu2021domain, lee2020biobert} to incorporating knowledge graphs \citep{yasunaga2022deep}. Subsequently, architectures such as GPT \citep{bubeck2023sparks} and Llama \citep{Touvron2023Llama2O} were trained on domain-specific \citep{wu2023pmc, gema2023parameter, yunxiang2023chatdoctor, zhang2023huatuogpt, labrak2024biomistral} or synthetic data \citep{tran2023bioinstruct, han2023medalpaca, kweon2023publicly}. Recent efforts have scaled up data and model parameters, resulting in GatorTronGPT \citep{peng2023study} and Clinical-Camel \citep{toma2023clinical}, along with studies on Flan-PaLM and PaLM-2's medical reasoning using chain-of-thought \citep{singhal2023large, singhal2023towards, wei2022chain, wang2022self}. Researchers have also introduced Meditron \citep{chen2023meditron}, an open-source medical LLM suite with 7B and 70B parameters, demonstrating superior performance against leading benchmarks and comparable results to GPT-3.5 and Med-PaLM-2. Despite the model size increase, performance gains have plateaued, and such expansions have not effectively addressed model forgetting \citep{wang2023survey, luo2023empirical}. In response, our study integrates a retriever with LLMs during training to tackle this issue, offering 7B and 13B versions. Our findings show that these models exceed traditional pre-trained language models in performance metrics, with our 13B model surpassing even the Meditron 70B.

\paragraph{Retrieval-Augmented Language Models} 
In the clinical domain, LLMs' factual inaccuracies due to gaps in medical knowledge pose significant risks, including misdiagnoses \cite{petroni2019language, sung2021can, yao2022context, singhal2023large}. 
Incorporating diverse knowledge repositories into LLMs enhances their performance across various NLP tasks, particularly in language modeling \citep{min2022nonparametric}. In traditional RAG models, the retrieval component is static, with updates applied only to the LLM. 
The process involves using the input as a query to first retrieve a collection of documents from a designated corpus, which the LLM then uses to enhance its predictions \citep{yu2022retrieval, izacard2022few}.
However, integrating retrieved documents does not guarantee effective use of external knowledge. Innovations like Atlas \citep{izacard2022few} and RETRO \citep{borgeaud2022improving} have modified models to better utilize retrieved information, but challenges remain in coherently blending this external data with the LLM’s pre-trained responses. 
Although some researchers have explored unsupervised training of retrievers to improve document relevancy for LLMs, such as REALM \citep{guu2020retrieval}, REPLUG \citep{shi2023replug}, and RA-DIT \citep{lin2023ra}, these systems are usually trained independently from LLMs. 
Our approach innovatively integrates and synchronously trains both the retriever and the LLM, ensuring the retriever supplies more appropriate documents, thereby boosting the LLM's effectiveness. 
We have validated our method's superiority through comparisons with step-by-step training techniques, confirming its enhanced performance in integrating external knowledge.

\section{Conclusion}
This study introduces the Joint Medical LLM and Retrieval Training (JMLR), significantly enhancing performance in medical question-answering and reasoning tasks. 
The JMLR models not only outperform existing state-of-the-art models in efficiently handling medical resources but also effectively reduce hallucinations in information generation, improving the accuracy and reliability of answers and explanations. 
Experimental results demonstrate substantial performance improvements across multiple medical benchmark tests, validating the effectiveness of integrating retriever and language model training. 
Furthermore, the reasoning capabilities of the JMLR model have been recognized by both GPT-4 and medical professionals, further confirming its potential and reliability for applications in the medical domain.

\section{Limitations and Ethical Considerations}
\label{sec:ethics}

This study offers valuable insights but also comes with several limitations that we would like to highlight:

\begin{itemize}
    \item \textbf{Domain Specificity:} Our research exclusively focuses on the task of medical QA and reasoning. The adaptation of the proposed method to other domains remains unexplored. This suggests that our approach may need further validation and adjustments before being applied to different fields.
    \item \textbf{Expertise of Annotators:} We relied on 3 doctors as annotators for human evaluation and preference results. While they are qualified to provide expert medical opinions and insights, employing more qualified domain experts as annotators would enhance the statistical significance of our results. We leave this to future work, along with addressing concerns about fairness, generalizability to other domains/languages, and potential biases inherent in LLMs.
\end{itemize}

\paragraph{Privacy Implications}
Despite significant advancements in the performance of medical knowledge acquisition and question-answering systems, privacy protection becomes a paramount concern when dealing with sensitive medical data. In particular, our model utilizes data from textbooks and public guideline. 

\paragraph{Bias Considerations}
Furthermore, while we strive to enhance the model's performance across multiple medical question-answering datasets, we must acknowledge that the choice and composition of datasets could introduce or exacerbate biases within the model. For example, if training data predominantly comes from certain geographic locations or populations, the model might exhibit biases towards medical conditions or treatment methods outside those groups. This could limit the fairness and effectiveness of the model when applied globally. Future research should consider training with more diverse and comprehensive datasets to reduce potential biases and enhance the model's universal applicability.

\paragraph{Broader Impacts}
Our study represents an important step forward in advancing AI applications in the field of medicine, but its broader societal impacts also require careful consideration. In particular, automated medical question-answering systems hold great potential in improving healthcare efficiency and accuracy but could also impact the roles of medical professionals and patient care practices. For instance, reliance on these systems may sometimes reduce direct communication between doctors and patients or might influence physicians' clinical judgment. Therefore, implementing these technological solutions should be approached with caution, ensuring they serve as a complement, not a replacement, to the toolkit of medical professionals. Moreover, the public's understanding and acceptance of these systems are crucial for their widespread use, necessitating enhanced education and transparency.

In summary, while our research demonstrates the potential of leveraging large language models and information retrieval techniques in medical question-answering systems, close attention must be paid to the ethical considerations of privacy, bias, and broader societal impacts. Future work should aim to address these challenges, ensuring the development and application of these technologies benefit the healthcare system and society as a whole.

\newpage

\bibliography{acl24.bib}
\bibliographystyle{acl_natbib}

\newpage

\appendix

\section{Appendix}
\label{sec:appendix}

\subsection{Case Study}


We presented four specific examples to further understand why JMLR outperforms other public models as shown in Table~\ref{tab:case1} and  Table ~\ref{tab:case2}. The first three examples are from the Amboss dataset. Due to privacy concerns with Amboss, we cannot display the complete content of these questions. The last example is from a USMLE question. Our answers for the first two questions are correct, while GPT-3.5 responded incorrectly. Although both our model and GPT-3.5 answered the last two questions correctly, there were some differences in the rationale generated. For the first example, we found that JMLR's retrieval function could extract similar questions, leading the model to answer incorrectly regarding direction. The retrieved document served as a background, fully utilizing the previously fine-tuned dataset. In the second example, we noticed that GPT-3.5 made a common-sense error: vaccination is not recommended for pregnant women in such cases. Even though GPT-3.5 has learned far more medical documents than our model by using pretraining or finetuning, it does not mean that the model will apply all medical knowledge correctly, even for some basic medical issues. In the third example, both we and GPT-3.5 answered correctly, but the rationale generated by GPT-3.5 differed significantly from the correct answer's rationale: GPT-3.5 simply stated some facts without providing a logical reason. However, our model gave a more detailed and logical explanation of why there are decreased circulating T cells, as JMLR could retrieve related documents, such as some documents about SCID, to better understand the underlying principles. The last example differs from the first three, as the highest-scoring document retrieved was from public guideline. The guideline provides a detailed introduction to PCOS and its symptoms, helping the model better explain. In contrast, GPT-3.5 simply correlated the symptoms straightforwardly without thoroughly analyzing other related symptoms, which can often lead to diagnostic errors, even though it answered this question correctly. Our model, however, performed a comprehensive analysis.

\subsection{Accuracy Variation with Different Numbers of Documents}

\begin{figure*}[!ht]
    \centering
    \includegraphics[width=\textwidth]{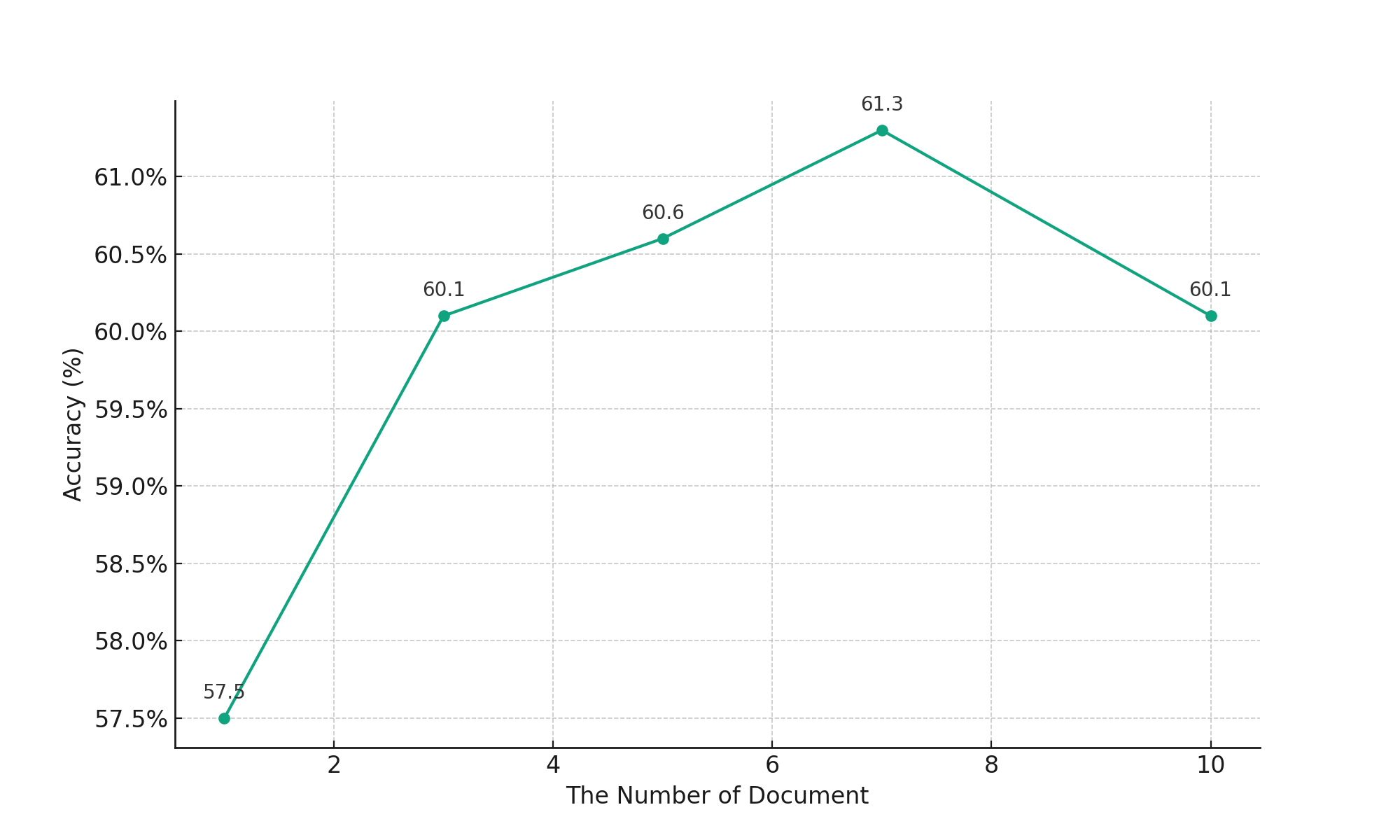}
    \caption{The horizontal axis represents the number of documents retrieved in JMLR, while the vertical axis shows the accuracy of JMLR tested on the USMLE.}
    \label{fig:doc}
\end{figure*}

During training, we set the number of background documents to seven, maintaining this count during the inference phase as well. This quantity is optimal, according to our experiments, as illustrated in the figure~\ref{fig:doc}. We employed the JMLR method to train the LLM on the MedQA dataset, retrieving varying numbers of documents. The performance is at its weakest when only one document is retrieved; this insufficiency leads to a lack of adequate medical knowledge for the model. As we increase the number of retrieved documents, the model's performance gradually improves, reaching its peak with seven documents. However, once this number increases to ten, performance again declines due to retrieving an excess of irrelevant documents, which hinders the model's ability to answer questions effectively.

\begin{table*}[!ht]
    \begin{scriptsize}
    \setlength{\tabcolsep}{2pt}
    \centering
    \begin{tabularx}{\textwidth}{lXXX}
    \toprule
    
    \multirow{2}{*}{1} 
    & 
    \textbf{Q1: A 48-year-old man...} \\
    & \textbf{Answer:}\\
    &
    \textcolor{red}{GPT-3.5:}

    The correct answer is A) Cerebral edema.

    Explanation:
    This patient presents with symptoms consistent with alcohol withdrawal syndrome (AWS), which can manifest as seizures (alcohol withdrawal seizures) due to abrupt cessation or reduction of alcohol intake in a chronic heavy drinker. AWS seizures typically occur within 48 hours of reducing alcohol intake.

    This patient's laboratory findings are indicative of hyponatremia (low serum sodium), which is common in severe alcohol withdrawal due to the syndrome of inappropriate antidiuretic hormone secretion (SIADH). SIADH leads to water retention and dilutional hyponatremia.

    The risk of cerebral edema is increased in patients with severe hyponatremia who are rapidly corrected to normal serum sodium levels. Rapid correction of hyponatremia can lead to osmotic shifts of water into the brain cells, causing cerebral edema, which may result in neurological deterioration, seizures, coma, and death. This risk is particularly high in chronic alcoholics due to adaptive changes in brain osmolytes.

    Given the patient's current status, urgent treatment to normalize serum sodium levels (such as with hypertonic saline infusion) puts him at increased risk for cerebral edema. Therefore, careful monitoring and slow correction of hyponatremia are essential in this patient to avoid this potentially fatal complication. \\
    & \textcolor{green}{Our:}

    This patient's presentation is consistent with alcohol withdrawal syndrome (AWS), which is characterized by autonomic hyperactivity, tremulousness, and seizures. AWS can occur in patients with a history of chronic alcohol use disorder who abruptly stop drinking or significantly reduce their alcohol intake. The syndrome is thought to be due to the withdrawal of the inhibitory effects of alcohol on the central nervous system.

    The most common complication of AWS is seizures, which can occur within 48 hours of the last drink. Seizures are often generalized tonic-clonic seizures and can be life-threatening. Other complications include delirium tremens (DTs), which is characterized by hallucinations, confusion, and autonomic instability. DTs can also lead to seizures and death.

    The treatment of AWS involves supportive care, including hydration, benzodiazepines, and possibly antipsychotics. In severe cases, hospitalization may be necessary for close monitoring and management of complications.

    Osmotic myelinolysis (also known as central pontine myelinolysis) is a condition that can occur in patients with severe hyponatremia (low sodium levels in the blood) who are treated with rapid correction of their sodium levels. This condition is characterized by demyelination of the central nervous system, leading to neurologic symptoms such as confusion, seizures, and coma.

    Given the patient's presentation and history of chronic alcohol use disorder, the most appropriate treatment is supportive care for AWS, including hydration, benzodiazepines, and possibly antipsychotics. Rapid correction of his sodium levels is not indicated due to the risk of osmotic myelinolysis. \\
    & \cellcolor{gray!10} \textbf{Expertise Comment: } \emph{GPT-3.5 incorrectly interpreted the patient's symptoms as primarily related to Alcohol Withdrawal Syndrome (AWS) without adequately considering the patient's laboratory results, particularly the severe hyponatremia (serum sodium 112 mEq/L). Although seizures can indeed accompany AWS, in this case, the diagnosis and treatment of hyponatremia should be the primary focus. Moreover, it incorrectly associated the risks of rapid correction of hyponatremia with cerebral edema. In fact, the main risk of rapidly correcting hyponatremia is osmotic myelinolysis, not cerebral edema. Osmotic myelinolysis results from rapid changes in blood osmolarity, leading to damage to the myelin sheath of brain neurons.}\\

    & \textcolor{brown}{\textbf{Retrieve:} A 48-year-old man is brought to the emergency department by his wife 20 minutes after she witnessed him vigorously shaking for about 1 minute. During this episode, he urinated on himself. He feels drowsy and has nausea. He has a history of chronic alcoholism; he has been drinking 15 beers daily for the past 3 days. Before this time, he drank 8 beers daily. His last drink was 2 hours ago. He appears lethargic. His vital signs are within normal limits. Physical and neurologic examinations show no other abnormalities. On mental status examination, he is confused and not oriented to time. Laboratory studies show: Hematocrit 44.0\% Leukocyte count 12,000/mm3 Platelet count 320,000/mm3 Serum Na+ 112 mEq/L Cl- 75 mEq/L K+ 3.8 mEq/L HCO3- 13 mEq/L Urea nitrogen 6 mEq/L Creatinine 0.6 mg/dL Albumin 2.1 g/dL Glucose 80 mg/dL Urgent treatment for this patient's current condition puts him at increased risk for which of the following adverse events?\" A: Cerebral edema B: Cardiac arrhythmia C: Hyperglycemia D: Osmotic myelinolysis E: Wernicke encephalopathy Correct Answer: D} \\

    \midrule \\
    & \textbf{Q2: A 27-year-old gravida 2, para 1 comes...} \\
    & \textbf{Answer:}\\
    & \textcolor{red}{GPT-3.5:}
    
    Based on the clinical presentation described, the most likely diagnosis for this patient is Rubella (German measles) infection during pregnancy, which can lead to congenital rubella syndrome (CRS) in the fetus if contracted during the first trimester. The typical symptoms of rubella include a maculopapular rash starting on the face and spreading downwards, along with lymphadenopathy and joint swelling.

    Given the risk of congenital rubella syndrome to the fetus, the most appropriate next step in management is:

    C) Vaccination

    Vaccination against rubella is contraindicated during pregnancy due to the risk of causing CRS. However, after delivery, the patient should receive the rubella vaccine to prevent future infections. This will not only protect the patient from future rubella infections but also prevent the risk of transmitting the virus to others, including vulnerable populations such as pregnant women.

    Options A, D, and E are not appropriate in this scenario. Oral acyclovir therapy (Option A) is used for herpes simplex virus infections, not rubella. Reassurance and symptomatic therapy (Option D) may provide comfort but do not address the potential risk to the fetus. Intravenous immunoglobulin therapy (Option E) is not indicated for rubella infection during pregnancy. Termination of pregnancy (Option B) is not necessary unless there are severe fetal abnormalities detected on further evaluation. \\
    & \textcolor{green}{Our:}

    The patient's history of a rash that started on her face and progressed downwards, along with the presence of postauricular lymphadenopathy, is suggestive of rubella infection. Rubella is a highly contagious viral infection that is spread through respiratory droplets or direct contact with infected secretions. It is characterized by a prodrome of malaise, low-grade fever, and lymphadenopathy, followed by the appearance of a pink maculopapular rash that begins on the face and spreads downwards. The rash is usually itchy and lasts for 3-5 days.

    Rubella infection during pregnancy, especially during the first trimester, can have serious consequences for the developing fetus. It can cause congenital rubella syndrome (CRS), which is characterized by a wide range of birth defects, including cataracts, deafness, heart defects, and intellectual disability.

    The most appropriate next step in management for this patient is termination of pregnancy. This is because the risk of CRS is high if the fetus is exposed to rubella during the first trimester, and the risk of birth defects outweighs the potential benefits of continuing the pregnancy.

    Options A, B, C, and E are not appropriate in this case. Acyclovir is used to treat herpes simplex virus infections, not rubella. Vaccination is not recommended during pregnancy due to the risk of fetal infection. Reassurance and symptomatic therapy are not sufficient to prevent the risk of CRS. Intravenous immunoglobulin therapy is not indicated for rubella infection during pregnancy. \\

    & \cellcolor{gray!10} \textbf{Expertise Comment: }GPT-3.5 made a made a basic error; it's not appropriate to vaccinate during pregnancy. \\
    \bottomrule
    \end{tabularx}
    \end{scriptsize}

\caption{Case Study For Retrieval Documents.}
\label{tab:case1}
\end{table*}

\begin{table*}[!ht]
    \begin{scriptsize}
    \setlength{\tabcolsep}{2pt}
    \centering
    \begin{tabularx}{\textwidth}{lXXX}
    \toprule
    
    \multirow{2}{*}{1} 
    & 
    \textbf{Q4: A 1-year-old girl is brought to the physician...} \\
    & \textbf{Answer:}\\
    &
    \textcolor{red}{GPT-3.5:}

    Given the clinical presentation and findings described, this patient is most likely suffering from severe combined immunodeficiency (SCID), which is characterized by a deficiency in both T and B lymphocytes.
    The additional finding most likely to be present in this patient with SCID is:
    D) Decreased circulating T cells
    SCID is a primary immunodeficiency disorder characterized by profound defects in both cellular and humoral immunity. Patients with SCID typically have absent or markedly reduced T cell counts, as T cell development and function are severely impaired. This deficiency in T cells predisposes patients to recurrent, severe, and often life-threatening infections.
    While B cells may also be affected in SCID, the absence of T cells is the primary hallmark of the disease. Therefore, decreased circulating T cells (Option D) would be the most likely additional finding in this patient.
    Other options such as increased circulating IgE (Option B) or increased circulating neutrophils (Option F) are less specific to SCID and could be seen in a variety of other conditions.
     \\
     
    & \textcolor{green}{Our:}

    Correct Answer: D
    Decreased circulating T cells

    This patient's severe combined immunodeficiency (SCID) is most likely due to an autosomal recessive adenosine deaminase deficiency (the second most common cause of SCID), which results in disrupted purine metabolism and accumulation of metabolites (including deoxyadenosine and dATP) that are toxic to maturing lymphocytes. Accumulation of dATP also inhibits the function of ribonucleotide reductase, which is essential for the generation of deoxynucleotides, thereby impairing the proliferation of lymphocytes. These effects on lymphocytes proliferation and maturation lead to a decreased number of circulating T cells, B cells, and NK cells, resulting in both cellular and humoral immunodeficiency. Clinically, patients with SCID have no tonsils, no palpable lymph nodes, and a rudimentary thymus.  In contrast, X-linked SCID (the most common variant) is caused by a defect in the IL-2 receptor, which is required for T-cell proliferation. Thus, X-linked SCID manifests with a decrease in circulating T cells and a normal number of B cells. Congenital immunodeficiency disorders

    Decreased levels of circulating parathyroid hormone in association with immunodeficiency can be seen in patients with DiGeorge syndrome. This syndrome also manifests with thymic hypoplasia and, possibly, the absence of tonsils. However, other features of CATCH-22 (e.g., congenital heart disease, dysmorphic facial features, hypocalcemia) would also be expected. Increased deoxyadenosine concentration in both the serum and urine and absent tonsils are not associated with DiGeorge syndrome.
    Increased circulating IgE levels can be seen in patients with Job syndrome, which is an immunodeficiency characterized by impaired neutrophil recruitment to the site of infection. Further manifestations would typically include coarse facial features, eosinophilia, severe eczema, and recurrent abscess formation. Increased deoxyadenosine concentration in both the serum and urine, absent tonsils, and an absent thymus shadow on x-ray are not associated with Job syndrome.
    Circulating B cells are decreased (not increased) in SCID due to adenosine deaminase deficiency and in X-linked agammaglobulinemia, an immunodeficiency characterized by a defective Bruton tyrosine kinase gene. Although X-linked agammaglobulinemia can also present with recurrent, severe pyogenic infections beginning > 4–6 months after birth, it is extremely rare in girls given its mode of inheritance. Absent thymic shadow, tonsillar hypoplasia, and elevated deoxyadenosine levels are also not consistent with this diagnosis.
    In X-linked SCID (the most common variant), the number of circulating B cells is normal. This patient's severe combined immunodeficiency (SCID) is most likely due to an autosomal recessive adenosine deaminase deficiency (the second most common cause of SCID), which results in disrupted purine metabolism and accumulation of metabolites (including deoxyadenosine and dATP) that are toxic to maturing lymphocytes. Accumulation of dATP also inhibits the function of ribonucleotide reductase, which is essential for the generation of deoxynucleotides, thereby impairing the proliferation of lymphocytes. These effects on lymphocytes proliferation and maturation lead to a decreased number of circulating T cells, B cells, and NK cells, resulting in both cellular and humoral immunodeficiency. Clinically, patients with SCID have no tonsils, no palpable lymph nodes, and a rudimentary thymus.  In contrast, X-linked SCID (the most common variant) is caused by a defect in the IL-2 receptor, which is required for T
    \\
    & \cellcolor{gray!10} \textbf{Expertise Comment: }GPT-3.5's explanation of the pathology is unclear, such as the specific type of SCID (Severe Combined Immunodeficiency caused by adenosine deaminase deficiency) and how this condition affects the proliferation and maturation of lymphocytes, leading to a reduced number of T cells. \\

    & \textcolor{brown}{\textbf{Retrieve:} ...The treatment for mild to moderate cases is mainly supportive, with early treatment of infections and prophylaxis with trimethoprim-sulfamethoxazole if needed, but severe cases require definitive therapy with hematopoietic stem cell transplantation. Congenital immunodeficiency disorders A defect in the interleukin-2 receptor subunit gamma gene (IL2RG) is the most common mutation associated with severe combined immunodeficiency (SCID), which can manifest soon after birth with recurrent episodes of otitis media, candidiasis, and sepsis. However, patients would also typically show failure to thrive and have a history of chronic diarrhea. A CBC, moreover, would reveal a low absolute lymphocyte count (< 1000/mm3). Delayed umbilical cord separation would not be expected. A defect in Bruton tyrosine kinase (BTK) is responsible for X-linked agammaglobulinemia (XLA), which may result in recurrent otitis media from infection with extracellular bacteria such as S. pneumoniae and H. influenzae. However, infants with B-cell defects such as XLA typically do not develop symptoms before 3\u20136 months because they generally retain passively acquired maternal antibodies up to this age. Delayed umbilical cord separation would also not be expected. Moreover, candidiasis is more commonly seen in defects involving T cells and granulocytes than in isolated primary defects of antibody production such as XLA. Defective NADPH oxidase results in chronic granulomatous disease (CGD), which may manifest in infancy with candidiasis (since C. albicans is catalase-positive) and neutrophilia during episodes of infection. However, these infections tend to remain localized, and sepsis is not as common as in other primary immune deficiency syndromes. Other pathognomonic features of CGD include GI obstruction and urinary retention caused by granulomas. Delayed umbilical cord separation would not be expected. Defective microtubules due to a loss of function of the lysosomal trafficking regulator gene (LYST) is the underlying pathophysiology of Chediak-Higashi syndrome (CHS), which can result in systemic candidiasis. However, patients with CHS would have neutropenia, not neutrophilia. Features such as oculocutaneous albinism and/or ocular motility disorders (e.g., strabismus, nystagmus) would also be present. Delayed umbilical cord separation would not be expected. A defectively organized actin cytoskeleton due to a mutation in the WAS gene is the underlying cause of Wiskott-Aldrich syndrome (WAS), which can manifest in infancy with recurrent otitis media as well as scaly erythematous lesions in the diaper area due to atopic dermatitis. However, the absence of thrombocytopenia on a CBC essentially rules out WAS. Delayed umbilical cord separation would also not be expected...} \\

    \bottomrule
    \end{tabularx}
    \end{scriptsize}
    \caption{Case Study For Retrieval Documents}
\label{tab:case2}
\end{table*}

\begin{table*}[!ht]
    \begin{scriptsize}
    \setlength{\tabcolsep}{2pt}
    \centering
    \begin{tabularx}{\textwidth}{lXXX}
    \toprule
    
    \multirow{2}{*}{1} 
    & 
    \textbf{Q3: A 27-year-old woman presents to the office with concerns about her long struggle with her physical appearance since adolescence. She says she has always been \"large\" and was constantly targeted by her classmates and coworkers for being so. Her main concern at the moment is her acne and unwanted facial hair on her upper lip, for which she often visits a local spa. She has tried numerous diet plans, exercise regimens, and cosmetic products with little to no effect. Recently, she underwent a glucose tolerance test that showed a plasma glucose level of 160 mg/dL (8.9 mmol/L) after 2 hours of a 75 g dose of oral glucose. She has a family history of type 2 diabetes mellitus and a menstrual cycle that occurs every 45 days. Her pulse is 72/min and the blood pressure is 138/80 mm Hg. On physical examination, her height is 160 cm (5 ft 3 in) and her weight is 85 kg (187 lb), and she has severe inflammatory acne over the cheeks and forehead and dark coarse hairs on the back. What is the most likely diagnosis of this patient? A: Hypothyroidism B: Idiopathic hirsutism C: Cushing's disease D: Polycystic ovarian syndrome (PCOS) E: Ovarian hyperthecosis} \\
    & \textbf{Answer:}\\
    &
    \textcolor{red}{GPT-3.5:}
    
    The most likely diagnosis for this patient is:
    D: Polycystic ovarian syndrome (PCOS)
    PCOS is a common endocrine disorder in women of reproductive age, characterized by hyperandrogenism, irregular menstrual cycles, and polycystic ovaries on ultrasound. The clinical features seen in this patient, including acne, unwanted facial hair (hirsutism), obesity, irregular menstrual cycles, and a family history of type 2 diabetes mellitus, are all consistent with PCOS.
    Other options:
    A: Hypothyroidism - While hypothyroidism can present with weight gain and menstrual irregularities, it is less likely given the presence of hirsutism and obesity in this patient.
    B: Idiopathic hirsutism - This is unlikely as idiopathic hirsutism typically presents with hirsutism without other features of PCOS such as irregular menstrual cycles and obesity.
    C: Cushing's disease - While Cushing's disease can present with weight gain and hirsutism, it is less likely in this patient as there are no signs of hypercortisolism such as central obesity, buffalo hump, or purple striae.
    E: Ovarian hyperthecosis - This is a rare condition characterized by luteinization of ovarian stromal cells leading to hyperandrogenism. However, PCOS is a more common cause of hirsutism and irregular menstrual cycles in young women.

     \\
     
    & \textcolor{green}{Our:}

    The patient's history of being "large" since adolescence, a family history of type 2 diabetes mellitus, and a menstrual cycle that occurs every 45 days, along with her physical examination findings of severe inflammatory acne over the cheeks and forehead and dark coarse hairs on the back, are suggestive of polycystic ovary syndrome (PCOS). PCOS is a common endocrine disorder among women of reproductive age and is characterized by hyperandrogenism, chronic anovulation, and polycystic ovaries on ultrasound. The condition is associated with insulin resistance, which can lead to hyperglycemia and impaired glucose tolerance, as seen in this patient's glucose tolerance test.

    The other options listed are not as likely given the patient's presentation:

    A: Hypothyroidism is characterized by a slowing of metabolic processes, including growth and development, and is not associated with the symptoms or findings seen in this patient.

    B: Idiopathic hirsutism is a condition characterized by excessive hair growth in women, often on the face, chest, and back, but it does not typically present with the other symptoms or findings seen in this patient.

    C: Cushing's disease is a condition caused by excessive production of cortisol, which can lead to weight gain, central obesity, and facial rounding, but it does not typically present with the other symptoms or findings seen in this patient.

    D: Polycystic ovarian syndrome (PCOS) is a common endocrine disorder among women of reproductive age and is characterized by hyperandrogenism, chronic anovulation, and polycystic ovaries on ultrasound. The condition is associated with insulin resistance, which can lead to hyperglycemia and impaired glucose tolerance, as seen in this patient's glucose tolerance test.

    E: Ovarian hyperthecosis is a condition characterized by excessive production of androgens by the ovaries, which can lead to hirsutism and acne, but it does not typically present with the other symptoms or findings seen in this patient.

    \\
    & \cellcolor{gray!10} \textbf{Expertise Comment: }Our explanation is slightly more detailed in terms of linking the patient's specific symptoms to the characteristics of PCOS and explaining why other diagnoses are less likely. It provides a more comprehensive understanding of PCOS and its relation to the patient's symptoms and test results. On the other hand, GPT-3.5 offers a more concise but still accurate diagnosis. \\

    \bottomrule
    \end{tabularx}
    \end{scriptsize}
\label{tab:example3}
\end{table*}

\subsection{Training Resource}
\label{apx:training_resource}
In our study, we utilized various training resources to evaluate the performance of different models on medical question-answering (QA) datasets. According to the data, for MedQA datasets, JMLR-13B, which requires 148 GPU hours for training, significantly outperforms Meditron-70B, which necessitates 42,630 GPU hours. In comparison, JMLR-7B needs approximately 100 GPU hours, RAG-7B requires 72 GPU hours, and RAG-13B needs 128 GPU hours.

The training time for JMLR models is considerably lower than that for Meditron-70B, though slightly higher than RAG-13B. However, JMLR models achieve state-of-the-art (SOTA) performance. Interestingly, we observe a substantial performance improvement in RAG-13B compared to RAG-7B. This improvement can be attributed to the fact that RAG models struggle to effectively learn how to understand or utilize medical documents. As a result, the performance of RAG-7B is limited by its model size. Conversely, RAG-13B, with its increased model size, is better able to comprehend documents, leading to enhanced performance. Our JMLR models, on the other hand, excel in enabling the model to understand and utilize medical documents effectively. This capability is reflected in their SOTA performance across multiple datasets, demonstrating the efficacy of our approach in medical QA tasks.

\subsection{Factuality metrics: UMLS-F1}
\label{apx:umls-f}
The assessment of factual accuracy in LLMs output leverages the UMLS concept overlap metric. The Unified Medical Language System (UMLS), established by~\cite{bodenreider2004unified}, significantly contributes to the biomedical domain's interoperability. It achieves this by amalgamating and disseminating a comprehensive collection of biomedical terminologies, classification systems, and coding standards from many sources. By doing so, UMLS aids in reconciling semantic variances and representational disparities found across different biomedical concept repositories.

For the identification and alignment of medical named entities within texts to their corresponding biomedical concepts in UMLS, we employed the Scispacy library~\footnote{We used the Scispacy \textit{en\_core\_sci\_lg} model.}.
Scispacy excels in identifying and clarifying entities, thus facilitating the accurate association of named entities found in LLMs output with the relevant UMLS concepts. 
This capability is critical for evaluating the LLMs output's factual accuracy.

The analytical process for LLMs output utilizes metrics of precision and recall. Precision represents the ratio of concepts present in both the LLM output and ground truth content, serving as a measure of the LLM output's factual correctness. In contrast, recall evaluates how well the information in the LLM output matches the intended content, reflecting the relevance of the presented information.

To calculate these metrics, we consider the concept sets from both the ground truth ($C_{ref}$) and the LLM output ($C_{gen}$). The formulas for recall and precision are as follows:

$$
\text{Recall} = \frac{|C_{ref} \cap C_{gen}|}{|C_{ref}|}
$$

$$
\text{Precision} = \frac{|C_{ref} \cap C_{gen}|}{|C_{gen}|}.
$$

The F1 score, derived from the above precision and recall values, is reported to provide a balanced measure of LLMs output's accuracy and relevance.

\subsection{Prompt for Evaluation}

For GPT-4 evaluation, we assessment each result from three ways: comprehension, reasoning step and recall of knowledge. Here is our prompt:

\begin{table*}

\label{tab:LLM-as-Medical-Student Output Examples for Physical Exams 2}
    \begin{scriptsize}
    \setlength{\tabcolsep}{5pt}
    \renewcommand{\arraystretch}{1.05}
    \centering
    \begin{tabularx}{\textwidth}{lp{12.1cm}}
    \toprule
    \multirow{2}{*}{GPT-4} & 
    \textbf{System}
    \newline
    ``\emph{Act as a USMLE evaluator, your role involves assessing and comparing a medical student's explanation to the provided target answer. Begin the assessment by carefully reviewing the provided target answer. Then, based on following specific criteria, determine the score for the student's answer.}'' \\
    & \textbf{Evaluation Criteria}
    \newline
    ``\emph{For each diagnosis, evaluate the medical student explanation base on the following three questions:}'' \\
    & \textbf{Question 1}
    \newline
    ``\emph{Does the medical student's answer contain any evidence of incorrect reading comprehension? (indication the question has not been understood)}'' \\
    & \textbf{Question 2}
    \newline
    ``\emph{Does the medical student's answer contain any evidence of incorrect reasoning steps? (incorrect rationale for answering the question)}'' \\
    & \textbf{Question 3}
    \newline
    ``\emph{Does the medical student's answer contain any evidence of incorrect recall of knowledge? (mention of an irrelevant and/or incorrect fact for answering the question)}'' \\
    & \textbf{Input}
    ``\emph{Medical student's answer: Reason generated by model}\\
    &``\emph{Medical student's answer: Reason generated by model} \\
    &``\emph{Target answer: Ground truth reason}\\
    &``\emph{Background Question: Question}\\
    & \textbf{Output Format}
    ``\emph{ Your evaluation should be provided in JSON format, as follows(don't generate any other information): \{\{"case 1": \{\{"question 1": "The score for question 1", "question 2": "The score for question 2", "question 3": "The score for question 3", \}\}, "case 2": "score for case 2 with the same format as case 1","case 3": "score for case 3 with the same format as case 1", "overall score": "the average score for question 1, 2, 3", "reason": "the reason why you give the score"\}\}}\\
    \bottomrule
    \end{tabularx}
    \end{scriptsize}
    \caption{Evaluation Prompt for GPT-4.}
    \label{gpt4-score-prompt}
\end{table*}

\begin{table*}[!ht]
\centering
\scalebox{0.8}{
\begin{tabular}{lllllll}
\toprule
& & MMLU & MedMCQA & MedQA & Amboss & Average \\
\midrule
\multirow{5}{*}{Non Corpus} 
& PMC-Llama-7B & 59.7 & 57.6 & 42.4 & 43.7 & 50.9 \\
& Llama 2-7B & 56.3 & 54.4 & 44.0 & 46.5 & 50.3 \\
& Meditron-7B & 55.6 & 59.2 & 47.9 & 50.1 & 53.2 \\
& ChatGPT & 69.4 & 51.0 & 50.2 & 49.1 & 54.9 \\
& Meditron70B & \textbf{73.6} & 65.1 & 60.7 & 76.4 & 68.9 \\
\midrule
\multirow{5}{*}{Open Guidelines} 
& RAG-7B & 63.8 & 62.1 & 47.3 & 50.7 & 55.9 \\
& RAG-13B & 69.8 & 63.4 & 56.8 & 60.7 & 62.7 \\
& RAG-ChatGPT & 68.9 & 55.2 & 53.3 & 49.3 & 56.7 \\
& JMLR-7B & 65.3 & 64.1 & 51.3 & 68.3 & 62.3 \\
& JMLR-13B & 70.1 & 64.5 & 59.5 & 79.6 & 68.4 \\
\midrule
\multirow{5}{*}{All Corpus} 
& RAG-7B & 62.1 & 62.4 & 54.6 & 70.7 & 62.5 \\
& RAG-13B & 69.9 & 64.2 & 59.9 & 76.9 & 67.7 \\
& JMLR-ChatGPT & 70.1 & 55.3 & 54.3 & 50.1 & 57.5 \\
& JMLR-7B & 64.3 & 64.2 & 56.2 & 71.2 & 64.0 \\
& JMLR-13B & 72.8 & \textbf{65.5} & \textbf{62.5} & \textbf{81.2} & \textbf{70.5} \\
\midrule
\multirow{1}{*}{MedRAG Corpus} 
& MedRAG-ChatGPT & 75.5 & 58.0 & 53.6 & 48.8 & 59.0 \\
\bottomrule
\end{tabular}
}
\caption{Since Meditron-70B has already been tested on MedQA, we are directly using its results~\citep{chen2023meditron}. For ChatGPT, we utilized the API of GPT-3.5-turbo to conduct tests on both MedQA and Amboss datasets. For RAG-7B and RAG-13B, we employed the same medical guidelines and medical QA bank that we used with JMLR.}
\label{tab:benchmark_scores}
\end{table*}


\end{document}